# SoRoSim

## A MATLAB Toolbox for Hybrid Rigid–Soft Robots Based on the Geometric Variable-Strain Approach

By Anup Teejo Mathew, Ikhlas Ben Hmida, Costanza Armanini, Frederic Boyer, and Federico Renda

Soft robotics has been a trending topic within the robotics community for almost two decades. However, available tools for the modeling and analysis of soft robots are still limited. This article introduces a user-friendly MATLAB toolbox, Soft Robot Simulator (SoRoSim), that integrates the geometric variable-strain (GVS) model of Cosserat rods to facilitate the static and dynamic analysis of soft, rigid, and hybrid robotic systems. We present a brief overview of the design and structure of the toolbox and validate it by comparing its results with those published in the literature. To highlight the toolbox's potential to efficiently model, simulate, optimize, and control various robotic systems, we demonstrate four sample applications. The demonstrated applications explore different actuator and external loading conditions of single-, branched-, open-, and closed-chain robotic systems. We think that the soft robotics research community will significantly benefit from the SoRoSim toolbox for a wide variety of applications.

## Introduction

One of the most trending topics in the robotics community is the development and design of soft robots that can tackle challenges otherwise hard and even impossible to solve using their traditional rigid counterparts [1]. Soft robots are lightweight, cheap, and adaptable to different environments and scenarios, as demonstrated by the vast number of applications where they have been employed. On the other side, their compliance and infinite number of degrees of freedom (DoF) intrinsically increase the complexity of their modeling.

Different modeling approaches have been proposed previously, varying in their simplifying assumptions and applicability. Some of the most commonly used approaches in soft robotics include the lumped mass model (LMM), finite-element (FEM)-based models, discrete elastic rod (DER) model, and the piecewise constant curvature (PCC) model. The LMM assumes soft links to be repeated segments of point masses connected by springs and dampers corresponding to their geometry, expected motion, and DoF. FEM-based models provide a way of numerically approximating partial differential equations governing the motion and deformation of the soft body. The PCC model constructs kinematic relations based on the robot's geometry and behavior under loading by discretizing its links into a finite number of circular arcs characterized by constant curvature [2]. Rod models, such as the Euler–Bernoulli beam, Timoshenko beam, and the Cosserat rod, model the material deformation of the robot or manipulator by assuming it to be a 1D continuum mechanics object. The development of precise theoretical models is crucial, but theory alone may not be enough to satisfy the demand for computational tools in the soft robotics field. Knowledge sharing initiatives become essential to allowing the growth of this relatively new research field [3]. Generalized modeling platforms eliminate the need to write robot-specific scripts for the simulation, design optimization, and model-based control of specific manipulators by providing user-friendly, accurate, fast, and reliable algorithms.

Popular examples of such platforms include, Soft Motion (SoMo) [4], a Python-based toolbox to simulate continuum manipulators as approximated spring mass systems, and Simulation Open Framework Architecture (SOFA) [5], a simulation tool that employs simplified FEMs. Toolboxes such as ChainQueen [6] and SimSOFT [7] also employ FEM-based modeling techniques to simulate soft robots. Titan [8], which is a GPU-accelerated C++ library, is another example of a simulator that models soft robots as a spring mass system. Toolboxes based on DERs include Elastica [9], which employs the Cosserat rod theory to model and control slender bodies with a finite number of lumped DoF, and Volume Invariant Position-based Elastic Rods (VIPER) [10], which uses a volume-invariant position-based elastic rods model to simulate the behavior of muscular hydrostats (muscle-like). Finally, the MATLAB package TMT Dynamics (TMTDyn) [11] uses discretized lumped systems and reduced-order models to control and analyze hybrid rigid–soft robots.

The majority of the available toolboxes for soft robotics modeling are based on FEM and lumped mass systems. These are theoretically simple but computationally heavy approaches, with a lot of nodes and DoF, designed for general-purpose simulation instead of analysis and control. Most of the DER-based simulators are oriented toward computer graphics rather than real mechanical systems. TMTDyn is a geometrically exact package based on the parametrization of positions and orientations rather than strains. Within this scenario, we present SoRoSim, which is a MATLAB toolbox based on the GVS approach and that directly extends rigid robot modeling techniques to soft and hybrid systems while maintaining a high level of accuracy. The toolbox takes advantage of the high fidelity of the Cosserat rod approach and the simplifying assumptions of the GVS approach that minimize the number of DoF required to represent the system and provide a geometrically exact framework, leading to accurate, fast, and computationally less expensive results. We implement a new computational approach based on a nested Gaussian quadrature scheme to solve the GVS formulation.

The SoRoSim toolbox allows the user to define and manipulate links (rigid and soft) and robotic systems (linkages) using user-friendly GUIs and the MATLAB workspace. The GUIs assist the definition of links, the links' assembly, the definition of the links' DoF, the assignment of constraints as closed-loop joints, and the application of external and actuation forces. A variety of typical external forces and actuation inputs are handled by GUIs, which provides a black-box experience, allowing users from all backgrounds to easily use the toolbox to perform static and dynamic analyses. The toolbox provides specific MATLAB files that the user can edit to input customized values of external and actuation forces as functions of joint coordinates, their derivatives, positions, velocities, and so on. Moreover, as a MATLAB toolbox, it can be used alongside built-in functions; add-ons, such as the Optimization Toolbox; and user-written codes to further facilitate the analysis and control of robotic systems. The intrinsic limitation of the toolbox is that it can model soft links only as Cosserat rods. However, rigid links can be modeled with no limitation on geometry and DoF. Figure 1 shows a small subset of robotic systems the toolbox can analyze. The SoRoSim toolbox package and user manual are available for free on [15]. The "Toolbox Design and Structure" section provides details on various aspects of SoRoSim toolbox, including its design, structure, and workflow.

We perform several toolbox sanity tests by comparing the toolbox's analysis results with published data and commercial software output. We compare the static equilibrium results with published solutions and ANSYS workbench results. We study the dynamic simulation of a flexible flying rod and compared it with existing literature. We also investigate the energy transfer among the kinetic, gravitational potential, and elastic potential energy of a cantilever beam under gravity. We discuss these validation studies in the "Toolbox Validation" section. In addition to validation, in the "Modeling Applications" and "Design Analysis and Control Applications" sections, we demonstrate four innovative applications where the soft robotics community can use the toolbox. The examples include the analysis of the static equilibrium and contact dynamics of hybrid robotic arms, an underwater locomotor, a design optimization problem, and a study of two cases of inverse dynamic control problems. Finally, in the "Discussion and Conclusion" section we discuss the computational performance of the toolbox and draw conclusions and future directions of SoRoSim.

## Governing Equations

The GVS approach was recently introduced by Renda et al. [16] in statics and Boyer et al. [17] in dynamics. It is based on a variable-strain parametrization of soft links represented by Cosserat rods, 1D slender rods that can bend, twist, stretch, and shear. Cosserat's model is the most general rod model, as it accounts for the orientation as well as the position of beam elements and allows for all six modes of deformation to be considered during analysis [2]. Since the strain is parameterized in the GVS approach, it is easy to disable any deformation modes. Thus, other beam theories can be kinematically reproduced. For instance, SoRoSim users can enable the rotational modes along with shear along $y$ and $z$ to create a Timoshenko beam. The GVS model is also geometrically exact and generalizes the geometric theory of rigid robotics to hybrid systems of soft and rigid links with multidimensional joints, externally applied point and distributed forces, and distributed actuation forces [18]. In this section, we give a summary of the model and an overview of the efficient computational techniques implemented in the SoRoSim toolbox.

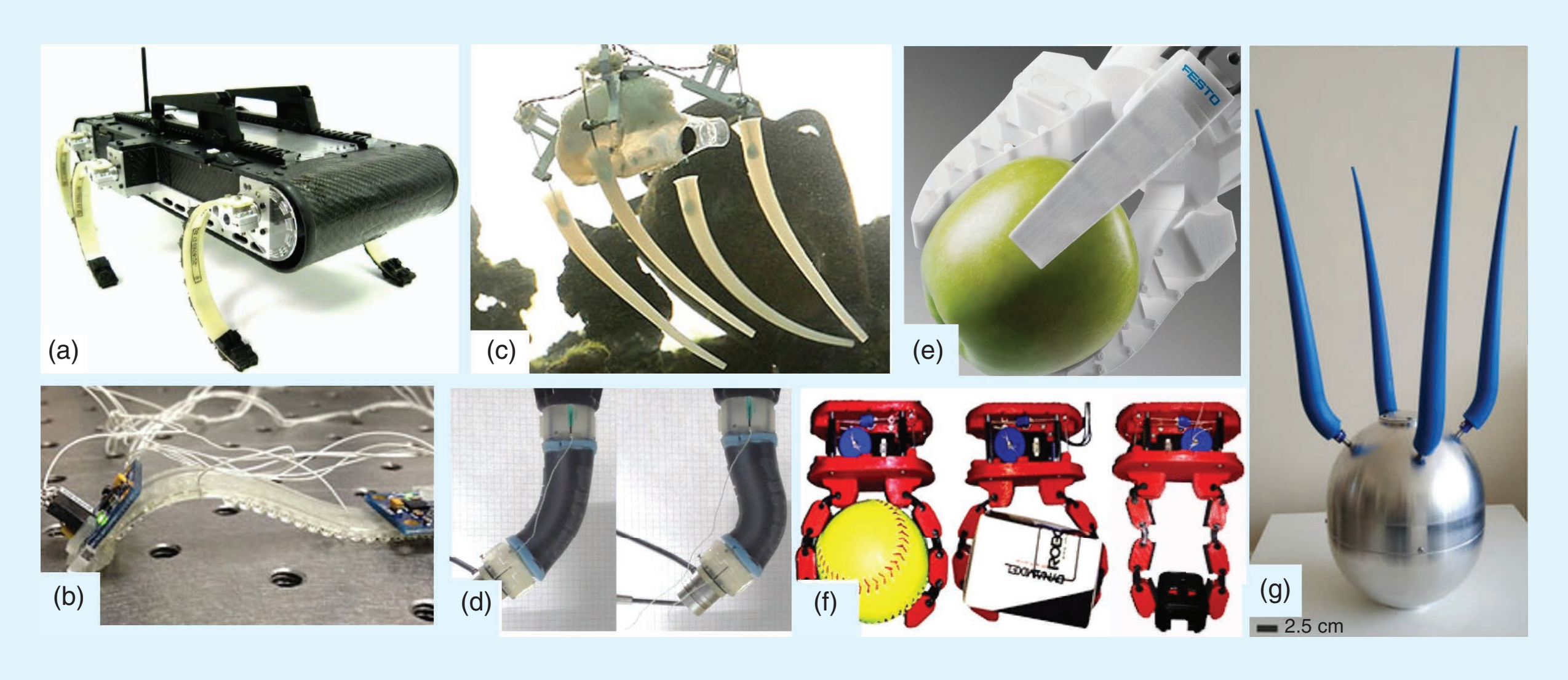

**Figure 1.** A visual representation of the various robotics systems the toolbox can model. These systems include rigid–soft hybrid robots, where the soft links can be approximated as Cosserat rods. (a) X-RHex, a hybrid robot with compliant legs. (b) Tuft Softworm, a soft robot that relies on shape memory alloy actuators and frictional forces for locomotion. (c) PoseiDRONE, a hybrid robot that uses tendon-based soft appendages to swim underwater and move on the seafloor [1]. (d) STIFF-FLOP, a pneumatic soft manipulator designed for use in surgical procedures [11]. (e) FinRay, a closed-chain soft gripper with rigid connectors [12]. (f) A two-finger tendon-driven hybrid gripper consisting of modular segments [13]. (g) A bioinspired hybrid flagellate robot for underwater applications [14]. The SoRoSim toolbox can model all these classes of robots by including contact, friction, and fluidic interactions as custom external forces similar to the examples in the "Modeling Applications" section.

Consider a floating hybrid kinematic chain composed of interconnected rigid and soft bodies represented by Cosserat rods. The configuration of a soft body $i$ (respectively, a rigid body) with respect to its predecessor in the chain is defined as a curve:

$$\boldsymbol{g}_i(\cdot): X_i \in [0, L_i] \mapsto \boldsymbol{g}_i(X_i) = \begin{pmatrix} \mathbf{R}_i & \mathbf{r}_i \\ \mathbf{0} & 1 \end{pmatrix} \in SE(3) \tag{1}$$

[respectively, a point $\boldsymbol{g}_i \in SE(3)$], mapping the body frame at $X_i$ to the body frame of the previous body at the reference configuration, as demonstrated in Figure 2.

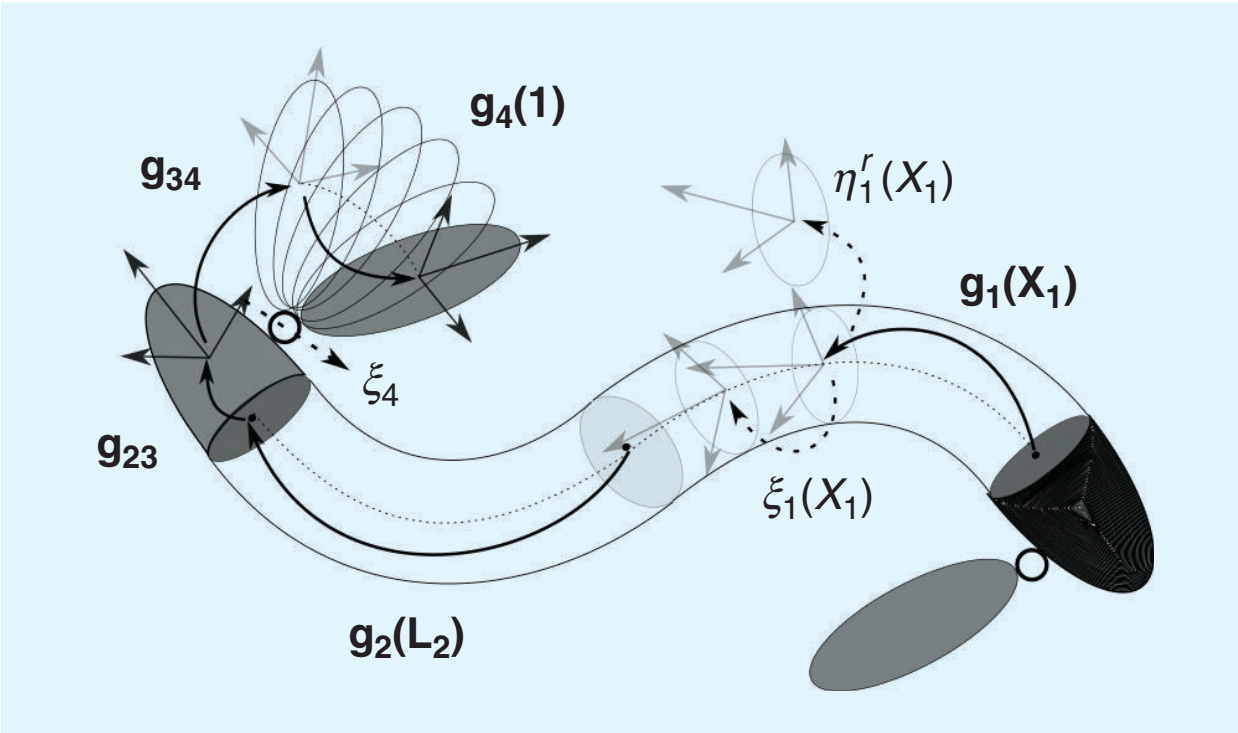

**Figure 2.** The proposed kinematics for a floating hybrid soft–rigid chain.

To study the exponential representation of $\boldsymbol{g}_i(X_i)$, we introduce its partial derivative with respect to space, $\boldsymbol{g}_i'(X_i) = \boldsymbol{g}_i\hat{\boldsymbol{\xi}}_i$, and with respect to time, $\dot{\boldsymbol{g}}_i(X_i) = \boldsymbol{g}_i\hat{\boldsymbol{\eta}}_i^r$, where $\boldsymbol{\xi}_i(X_i) \in \mathbb{R}^6$ defines the strain twist in the body frame, $\boldsymbol{\eta}_i^r(X_i) \in \mathbb{R}^6$ is the velocity twist relative to the predecessor in the body frame, and $\widehat{(\cdot)}$ is the isomorphism from $\mathbb{R}^6$ to $se(3)$. The space derivative of $\boldsymbol{g}_i(X_i)$ is a matrix differential equation that can be integrated in space according to $\boldsymbol{g}_i(X_i) = \exp(\hat{\boldsymbol{\Omega}}_i(X))$, where $\hat{\boldsymbol{\Omega}}$ is the Magnus expansion of the field $\hat{\boldsymbol{\xi}}$. For the rigid link case, $\hat{\boldsymbol{\xi}}_i(X_i)$ is constant in $X_i$ and equal to the body frame representation of the joint twist attached to $i$, while for the soft link case, $\hat{\boldsymbol{\xi}}_i(X_i)$ is variable. The equality of the mixed partial derivative of $\boldsymbol{g}_i$ provides the relation between the time derivative of the strain twists and the link's relative velocity and acceleration twists [19]:

$$\begin{aligned} \boldsymbol{\eta}_i^{r\prime}(X_i) &= \dot{\boldsymbol{\xi}}_i - \mathrm{ad}_{\boldsymbol{\xi}_i}\boldsymbol{\eta}_i^r, \\ \dot{\boldsymbol{\eta}}_i^{r\prime}(X_i) &= \ddot{\boldsymbol{\xi}}_i - \mathrm{ad}_{\dot{\boldsymbol{\xi}}_i}\boldsymbol{\eta}_i^r - \mathrm{ad}_{\boldsymbol{\xi}_i}\dot{\boldsymbol{\eta}}_i^r, \end{aligned} \tag{2}$$

where $\mathrm{ad}_{(\cdot)} \in \mathbb{R}^{6\times 6}$ is the adjoint operator of $\mathfrak{se}(3)$ [20].

It is time now to discretize the system and introduce the generalized coordinates. The continuous strain fields $\boldsymbol{\xi}_i(X_i)$ are parametrized by a finite functional basis of strain modes [16]:

$$\boldsymbol{\xi}_i(X_i) = \boldsymbol{\Phi}_{\xi_i}(X_i)\,\boldsymbol{q}_i + \boldsymbol{\xi}_i^*(X_i) \tag{3}$$

where $\boldsymbol{\Phi}_{\xi_i}(X_i) \in \mathbb{R}^{6\times n_i}$ ($n_i$ being the number of DoF of link $i$) is a matrix function whose columns form the basis for the strain field, $\boldsymbol{q}_i \in \mathbb{R}^{n_i}$ is the vector of coordinates in that basis, and $\boldsymbol{\xi}_i^*(X_i) \in \mathbb{R}^6$ is a reference strain whose primary function is to model nonzero yet constrained strains, such as inextensibility. Note that the matrix $\Phi_{\xi_i}(X_i)$ is constant for rigid joints.

The integration of (2) using (3) for all the bodies leads to the definition of the geometric Jacobian $[\boldsymbol{J}_i(\boldsymbol{q}, X) \in \mathbb{R}^{6\times n}]$ and its derivative $\dot{\boldsymbol{J}}_i(\boldsymbol{q}, \dot{\boldsymbol{q}}, X) \in \mathbb{R}^{6\times n}$ $(n = \Sigma n_i)$. Once $\boldsymbol{J}_i$ and $\dot{\boldsymbol{J}}_i$ are found, we project the free dynamics of the floating hybrid chain onto the space of generalized coordinates to yield the generalized dynamics of the system:

$$\boldsymbol{M}\ddot{\boldsymbol{q}} + (\boldsymbol{C} + \boldsymbol{D})\,\dot{\boldsymbol{q}} + \boldsymbol{K}\boldsymbol{q} = \boldsymbol{B}\boldsymbol{u} + \boldsymbol{F} \tag{4}$$

where $\boldsymbol{M}(\boldsymbol{q}) \in \mathbb{R}^{n\times n}$ is the mass matrix, $\boldsymbol{C}(\boldsymbol{q}, \dot{\boldsymbol{q}}) \in \mathbb{R}^{n\times n}$ is the Coriolis matrix, $\boldsymbol{D} \in \mathbb{R}^{n\times n}$ is the damping matrix, $\boldsymbol{K} \in \mathbb{R}^{n\times n}$ is the stiffness matrix, $\boldsymbol{B}(\boldsymbol{q}) \in \mathbb{R}^{n\times n_a}$ ($n_a$ being the total number of actuators) is the actuation matrix, $\boldsymbol{F}(\boldsymbol{q}, \dot{\boldsymbol{q}}) \in \mathbb{R}^n$ is the vector of generalized external forces, and $\boldsymbol{u} \in \mathbb{R}^{n_a}$ is the vector of applied actuation forces.

We developed a recursive two-level nested quadrature scheme to estimate the coefficients of (4). For soft links, the SoRoSim toolbox uses the Gauss quadrature numerical integration method (the order is chosen by the user) to evaluate these coefficients. Stiffness ($\boldsymbol{K}$) and damping ($\boldsymbol{D}$) coefficients, which are associated with linear elastic models, are precomputed offline. However, the framework also allows the computation of nonlinear strain-dependent constitutive laws. Apart from ($\boldsymbol{K}$) and ($\boldsymbol{D}$), all the other coefficients ($\boldsymbol{M}$, $\boldsymbol{C}$, $\boldsymbol{B}$, and $\boldsymbol{F}$) of (4) are computed as functions of $\boldsymbol{q}$ and $\dot{\boldsymbol{q}}$.

Estimating these coefficients involves assessing $\boldsymbol{g}_i$, $\boldsymbol{J}_i$, and $\dot{\boldsymbol{J}}_i$ at every evaluation point, such as Gaussian points of the soft divisions and the center of mass of rigid links. In our previous article, we used a fourth-order Zannah collocation approximation to estimate the value of $\boldsymbol{g}_i$ recursively: $\boldsymbol{g}_i(X_i + h_k) = \boldsymbol{g}_i(X_i)\exp(\hat{\boldsymbol{\Omega}}_i^k(h_k))$, where $h_k$ is the material length in the $k$th interval between two consecutive Gauss quadrature points and $\hat{\boldsymbol{\Omega}}_i^k(h_k)$ is the approximation of the Magnus expansion [16]. Inspired by this, we derived recursive formulations for the computation of $\boldsymbol{J}_i$ and $\dot{\boldsymbol{J}}_i$. The complete theory and description of the computational strategy behind the toolbox are available at the link provided in [15].

Equation (4) is an ordinary differential equation that could be solved using explicit time integrators, such as "ode45" and "ode15s" in MATLAB. The static equilibrium equation of the system can be derived from (4) by equating the time derivatives of $\boldsymbol{q}$ ($\dot{\boldsymbol{q}}$ and $\ddot{\boldsymbol{q}}$) to zero. The resulting static equation could be solved numerically using root finder functions, such as "fsolve" in MATLAB. For the case of closed-chain robots, additional terms corresponding to the constraint forces due to the closed-loop joints will be present in (4) [12].

## Toolbox Design and Structure

We developed the SoRoSim toolbox in MATLAB, which provides users with a vast library of functions for mathematical computations and gives access to various built-in functions, add-ons, and toolboxes to analyze different aspects of robotic systems. We also employ an object-oriented programming (OOP) approach, which entails program design around data and objects rather than functions and logic. OOP allows the developer to group "objects" with similar attributes under a "class," providing a well-structured map of the program and allowing easy access and adjustment to object-specific data ("properties") with the help of class-specific "methods." The SoRoSim toolbox consists of three MATLAB class elements: SorosimLink, Twist, and SorosimLinkage. These classes work together to facilitate linkage creation and simulation through a sequence of user-friendly GUIs.

The SorosimLink class allows the user to construct soft and rigid links with various joint types and geometry. The user can choose from nine different types of lumped joints (fixed, revolute, prismatic, helical, cylindrical, universal, planar, spherical, and free) and three default cross-sectional shapes (circular, rectangular, and ellipsoidal). However, analysis performed within the toolbox is not limited to the link's default cross-sectional shapes. Once a link is defined, the user can update its properties, such as the screw inertia matrix and stiffness matrix, to account for any arbitrary cross-sectional shape and nonhomogeneous mass distribution. The shape may also vary as a function of curvilinear abscissa along the link axis, $X$. The default material model used to compute the cross-sectional screw stiffness matrix is a linear elastic model. This material model provides an accurate representation of the material behavior when it is subjected to strains that do not exceed 100%; since this is the case for most soft robotics applications this is an appropriate material model to use. However, the toolbox allows users to use a custom material model by modifying the elasticity tensor matrix in the SorosimLink class. An overview of the SorosimLink creation is provided in Figure 3(a).

The Twist class specifies the active DoF of lumped joints and the deformation modes and corresponding strain orders of soft link divisions. For a soft link division, the class allows the user to enable the required modes among six deformation modes: torsion about the $x$-axis, bending about the $y$-axis, bending about the $z$-axis, elongation along the $x$-axis, shear along the $y$-axis, and shear along the $z$-axis. The order of a particular mode corresponds to the polynomial that is used to estimate the strain values. The Twist class also allows the user to define a reference strain value for the soft division corresponding to its rest configuration. The inset in Figure 3(b) shows a sample GUI with which the user creates the Twist class for a soft link division. The GVS model allows the definition of the strain base of the soft links as a continuous or discontinuous function of $X$. The user may change the default polynomial basis of soft links once the SorosimLinkage is defined.

The SorosimLinkage class allows the user to assemble previously defined links into various single-, branched-, open-, and closed-chain systems. SorosimLinkage calls the Twist class to determine each link's DoF. The user can also add closed-loop joints by selecting appropriate joint types. The class allows the definition of various external forces and actuation inputs. The class automatically precomputes and saves constant properties of the linkage, such as the generalized stiffness matrix (**K**) and the generalized damping matrix (**D**). It also allows the users to program custom external and actuation forces. An overview of the SorosimLinkage creation is given in Figure 3(b).

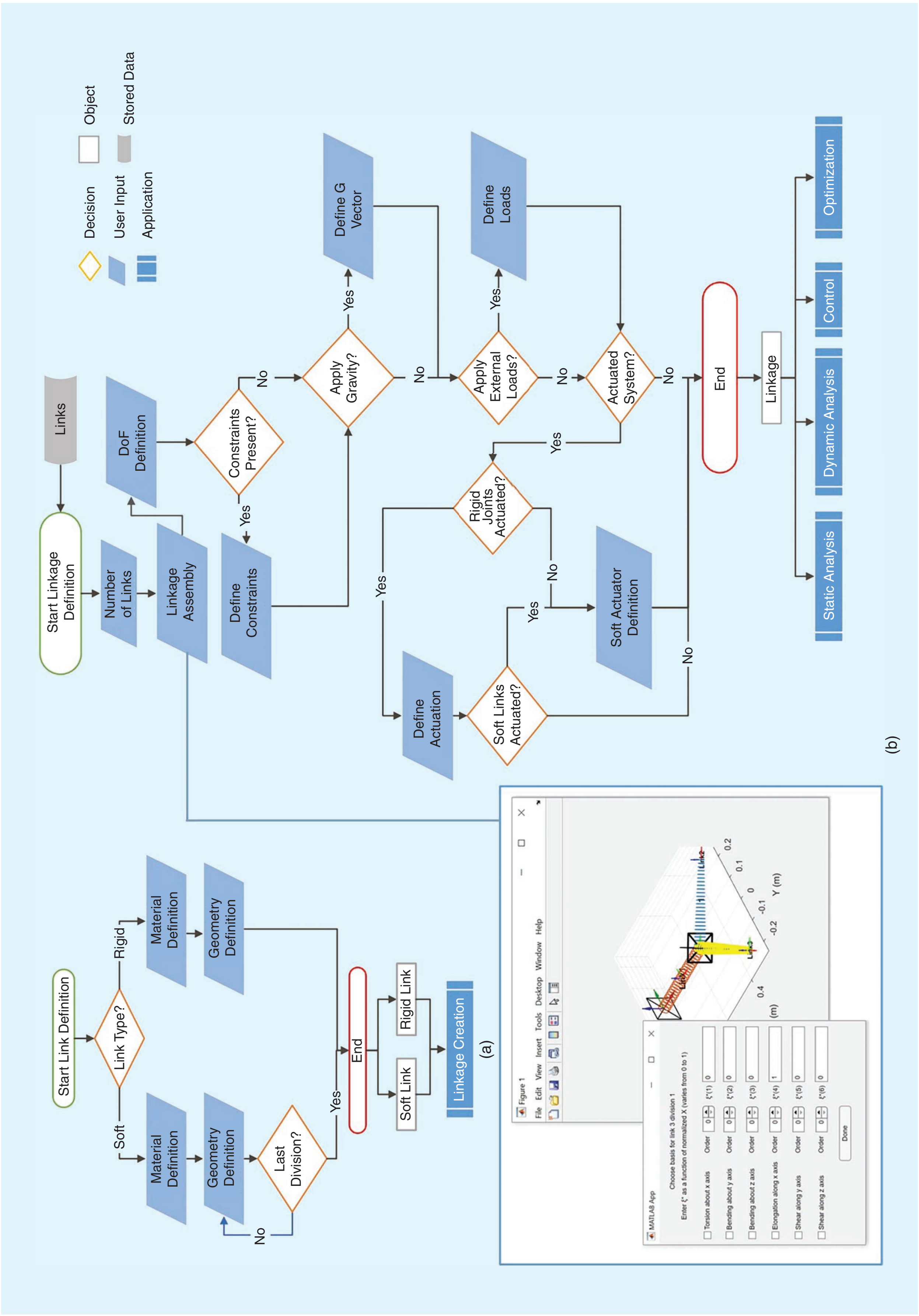

**Figure 3.** A toolbox overview: (a) the SorosimLink creation and (b) the SorosimLinkage creation and applications. Inset: a preview of the GUI.

We pack the SorosimLinkage class with "methods" that facilitate the analysis of multibody systems and the postprocessing of results (static equilibrium configuration and dynamic video output). The methods for analysis include functions to solve the GVS model for static and dynamic analyses. The SorosimLinkage methods can also compute values of system parameters, such as the Jacobian (**J**), generalized mass (**M**), and Coriolis (**C**) matrices for a given value of $\boldsymbol{q}$ and $\dot{\boldsymbol{q}}$. Users can utilize these methods for problem-specific analyses. The reader may refer to the toolbox manual [15] for a detailed description of all the properties and methods of SoRoSim classes.

## Toolbox Validation

To validate the toolbox, we conduct several numerical tests and comparisons with verified and published data. We present these tests in this section.

### Test 1: Fixed–Free Beam With a Follower Tip Force

We first simulate the bending behavior of a cantilever beam with a follower force applied at the tip. Many authors have considered this problem and solved it using numerical approaches, such as the finite-strain rod method [21] and other geometrically exact models [17]. Using the toolbox, we construct a 100-m-long cylindrical beam with Young's modulus $E = 6.75$ GPa and a diameter of 57 cm. We set a fourth-order bending strain about the $y$-axis to model the deflection of the rod when subjected to the follower tip force. The force is varied between 0 N and 130 kN. We use 15 Gauss quadrature points, specified during the SorosimLink creation process, for the computation of integrals (in this case, **K**). Figure 4(a) illustrates the deflection of the link as well as the horizontal and vertical displacement of the tip as modeled by the toolbox. The results obtained match those obtained in [21], as detailed in Figure 4(b). The total DoF of the rod modeled using SoRoSim is 5 DoF, while Simo et al. [21] used a 1D finite-element mesh consisting of five elements with quadratic shape functions corresponding to 10 DoF. This demonstrates the GVS approach's ability to recreate accurate results by using fewer DoF. Additionally, the reported computation time is 16.4 s per loading step in [17], while it took 42 ms for the same in the SoRoSim toolbox.

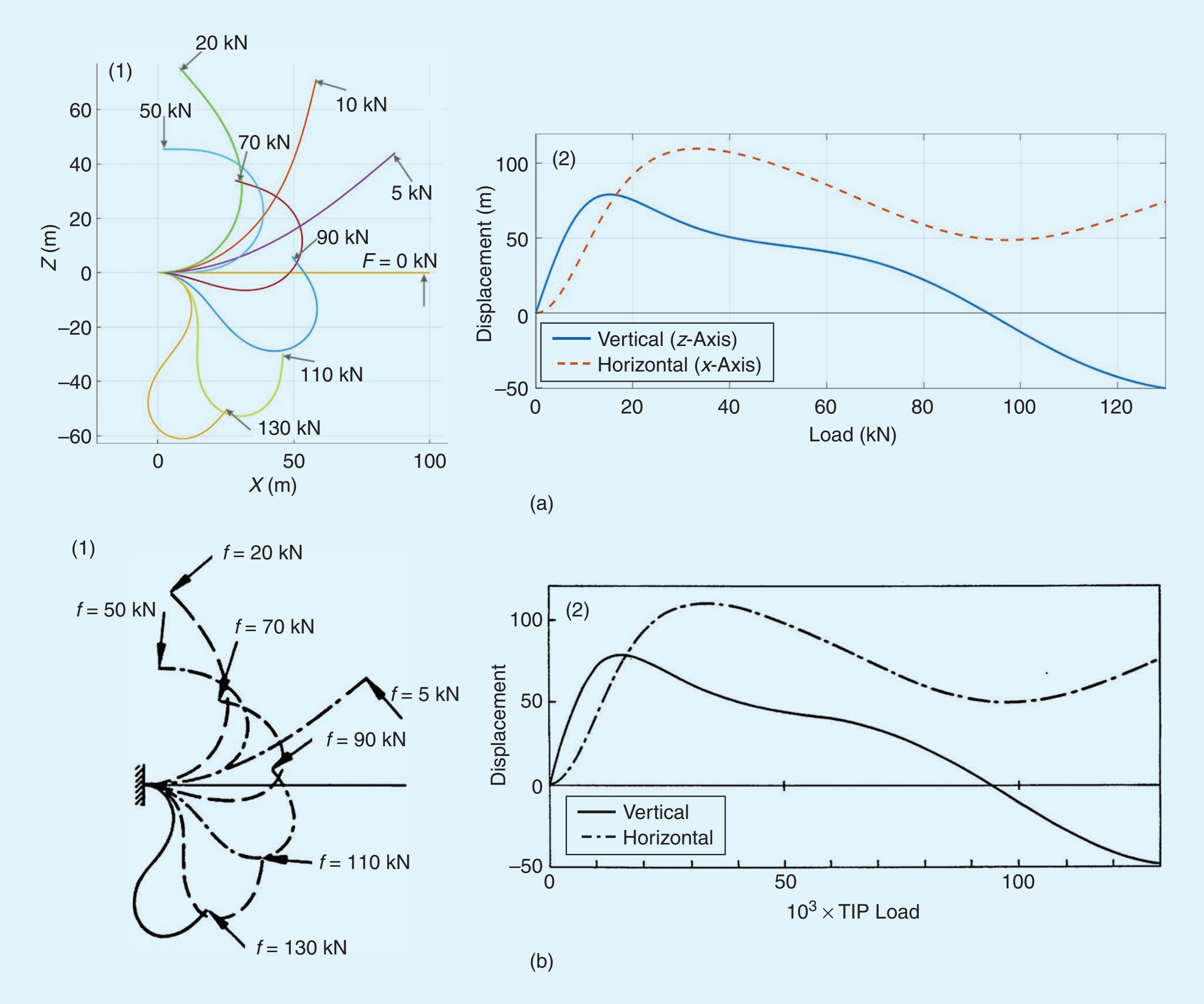

**Figure 4.** (a) The toolbox simulation output. (b) The results obtained in [21]. (1) The clamped beam profile under varying follower tip loads. (2) The horizontal and vertical tip displacement at different loads.

We also use this example to highlight the scaling process the toolbox uses. The toolbox performs internal computations on a soft division after normalizing its length into one unit (here, there are 100 m to one unit). Consequently, physical quantities with length dimensions are scaled using the original length of the division. Once the simulation is complete, the toolbox scales back the resulting values of joint coordinates into their actual dimensions. The normalization of soft divisions avoids poorly scaled matrices, such as the generalized stiffness matrix. This allows faster static solutions and more stable dynamic simulations.

### Test 2: FEM Study of a Fixed–Fixed L-Shaped Beam

To test the performance of the toolbox with respect to commonly used methods of modeling, we compare the results obtained when a soft linkage is subjected to a distributed load (gravity) with those obtained through FEM simulation. Two 0.7-m-long soft links with a $5 \times 5$-cm square cross section (aspect ratio: 14:1) are connected to form an L-shaped linkage clamped at each end. We use a fixed closed-loop joint to fix the rear end of the second link with the ground. All six deformation modes are enabled for this simulation; quadratic and linear polynomials are used to estimate the rotational and translational modes, respectively. Hence, there are 15 DoF for a link (30 in total). For the material, $E = 10$ MPa; Poisson's ratio, $\nu = 0.5$; and density, $\rho = 1,200$ kg/m$^3$, are used.

We compare the static equilibrium results obtained from the toolbox with those of ANSYS Workbench. A linkage with the same geometry and material properties is created in ANSYS. A total of 152 quadratic 3D elements with 1,062 nodes (~3,000 DoF) are used for the simulation. Figure 5(a) describes the toolbox result, while Figure 5(b) conveys that of ANSYS. The maximum deformation obtained from the toolbox result is 12.98 cm, whereas the FEM simulation gives a deformation of 13.27 cm. Hence, with fewer DoF (1:100), the GVS method can estimate the deformed shape of the L-shaped beam. The simulation results also suggest that the toolbox can effectively handle closed-loop problems.

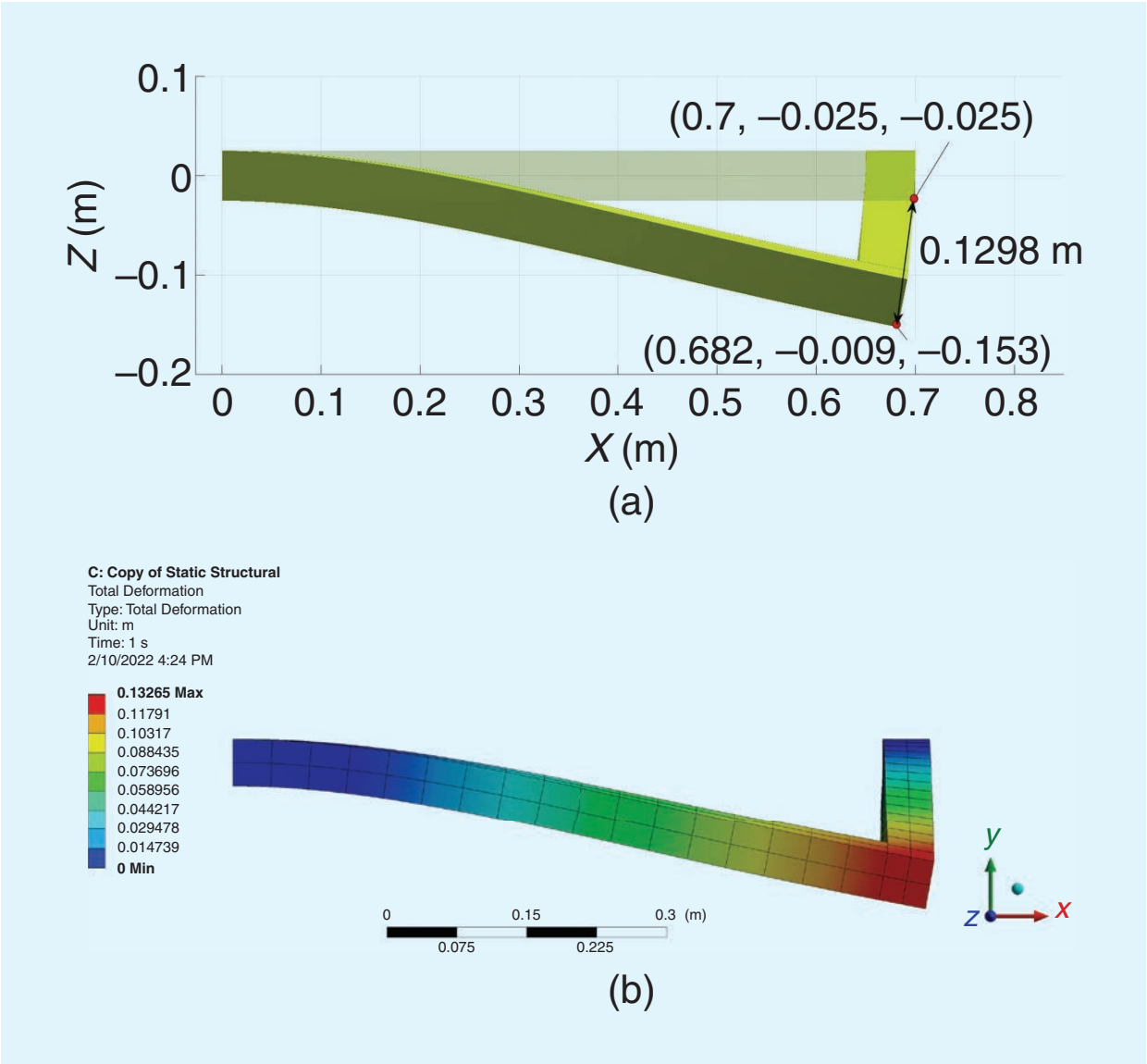

**Figure 5.** (a) The SoRoSim results showing the reference and deformed shape of the linkage. (b) The total deformation and deformed shape obtained from ANSYS.

### Test 3: Dynamics of a Flexible Flying Rod

In this comparison, we look at the dynamics of a freely flying flexible rod (also known as the *flying spaghetti problem*) which is a problem introduced by Simo and Vu-Quoc [22] and replicated in [17]. We model a 10-m-long soft rod, with a free lumped joint and initially at rest in the position shown in Figure 6(a)(1). The position and orientation of the base of the soft rod are parameterized by lumped DoF of the free joint. We used the inextensible Kirchhoff model with a quadratic polynomial basis to define the basis of the rod. Hence, including the six DoF of the free joint, there are 15 DoF in this system. Time-dependent point force $F1$ and moments $M2$ and $M3$ are applied at the tip of the rod, as in Figure 6(a)(1). The magnitude of $M2$ is defined as a triangular pulse function that starts at time $t = 0$ s, peaks at 200 N·m in 2.5 s, and goes back to zero at 5 s. The numerical values of the magnitude of $F1$ and $M3$ are 1/10 and half of $M2$, respectively. The user can define such dynamic inputs as a function of time ($t$) in the GUI.

We perform the dynamic analysis of the system for the first 7 s. Figure 6(a)(2) and (3) show two views of superimposed snapshots of the rod in midflight at different times, as solved by the SoRoSim Toolbox. The rod's position, orientation, and deformation match exactly with the published results in [17], provided in Figure 6(b)(2) and (3). This example also demonstrates the capabilities of the toolbox in modeling lumped and distributed joints (soft body) within the same framework.

We use this example to highlight the computational efficiency of the toolbox. The system uses 15 DoF to simulate a complex dynamic motion in 3D. Boyer et al. report a computational time of 4 h, 30 min for a 30-s simulation of the same problem [17]. Hence, on average, the reported computational time for a second of the simulation was 9 min. Using SoRoSim, we were able to solve a 7-s simulation in less than 1 s of computational time, which is three orders ($> 3,000\times$) faster than the previously reported computational time. Faster-than-real-time computation ($7\times$ faster in this case) will allow the use of SoRoSim for real-world applications. To better appreciate the dynamic simulations presented in this article, the reader may refer to the supplementary video available at https://doi.org/10.1109/MRA.2022.3202488.

### Test 4: Energy Balance in SoRoSim Dynamics

Here, we investigate the energy balance in damped and undamped cantilever beams released from rest with an initial strain under gravity. We create a soft link with a length ($L$) of 0.5 m and a radius that linearly decreases from 2 to 1 cm. The material properties of the link are $E = 1$ MPa, $\nu = 0.5$, and $\rho = 1{,}000$ kg/m$^3$. For the damped rod, we use an elastic damping of 11.2 KPa. We enable all three angular modes of deformation (torsion and rotations about the $y$- and $z$-axes) with a cubic polynomial approximation for the strains. To obtain a complex dynamic motion, the soft link is released from rest with an initial bending of 1 rad/m about the $y$-axis. We run the dynamic simulation for 5 s. Figure 7 describes the motion of the damped and the undamped rod between time $t=0$ and $t=0.75$ s.

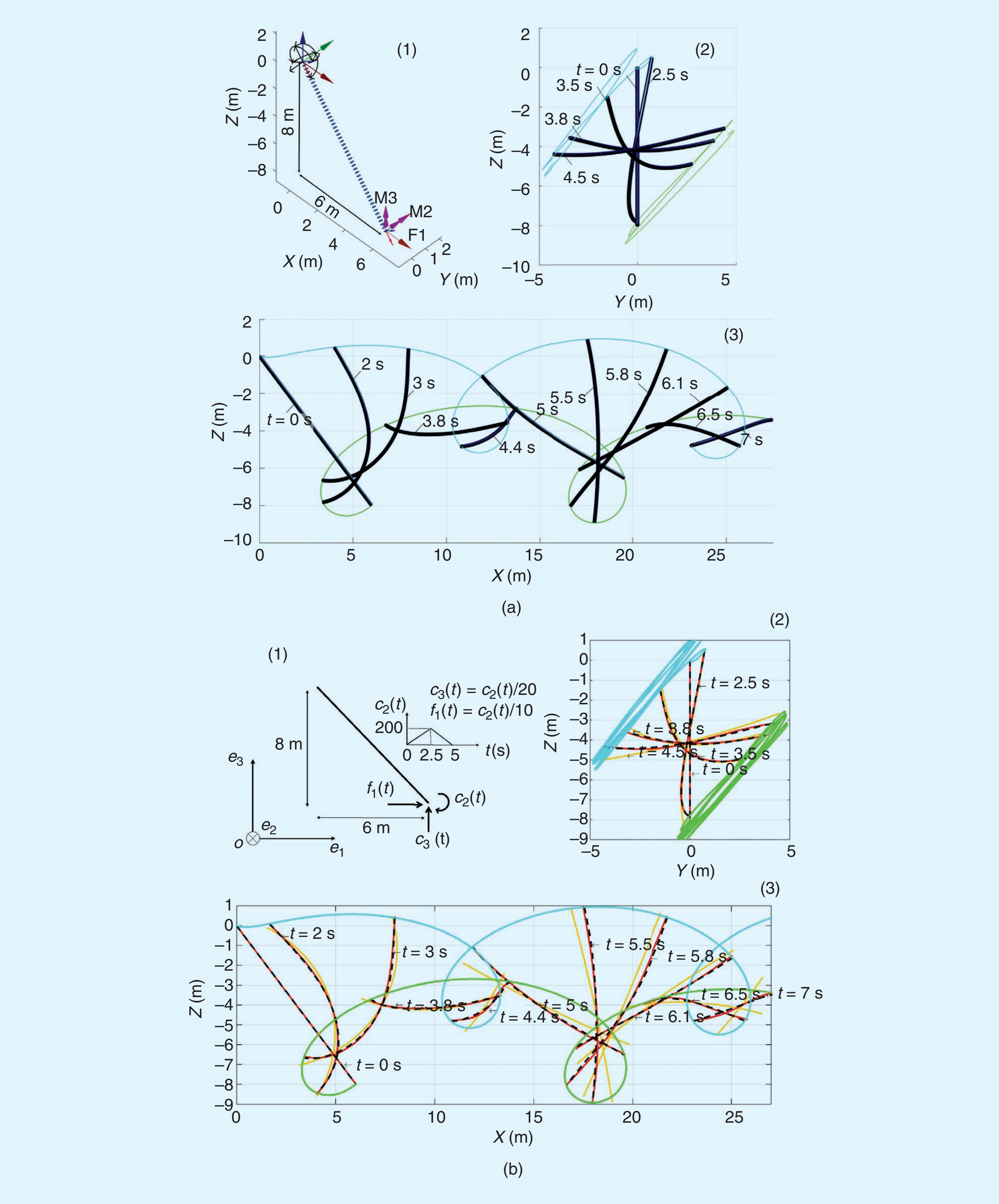

**Figure 6.** A comparison of (a) the simulation results obtained using SoRoSim with (b) those obtained in [17]. (1) The reference configuration of the flexible rod with external force and moments. Snapshots of the midflight dynamics of the rod in (2) the $yz$-plane and (3) the $xz$-plane. The comparison is with the dashed black line, which models an inextensible rod with three modes/strain.

The total energy of the beam at a given time is given by the sum of kinetic, gravitational potential, and elastic potential energies:

$$\begin{aligned} E_{tot} &= \frac{1}{2}\int_0^L \boldsymbol{\eta}^T \bar{\mathcal{M}}\boldsymbol{\eta}\, dX + \int_0^L \rho.A.g.h dX \\ &\quad + \frac{1}{2}\int_0^L (\boldsymbol{\xi}-\boldsymbol{\xi}^*)^T \boldsymbol{\Sigma}(\boldsymbol{\xi}-\boldsymbol{\xi}^*)\, dX \\ &= \frac{1}{2}\dot{\boldsymbol{q}}^T \boldsymbol{M}\dot{\boldsymbol{q}} + U_g(\boldsymbol{q}) + \frac{1}{2}\boldsymbol{q}^T \boldsymbol{K}\boldsymbol{q} \end{aligned} \tag{5}$$

where $\boldsymbol{\eta}$ is the screw velocity, $\bar{\mathcal{M}}$ is the cross-sectional screw inertial matrix, $A$ is the area of cross section, $g$ is acceleration due to gravity, $h$ is the $y$ coordinate of a cross section of the rod (in the opposite direction of gravity), $\boldsymbol{\xi}$ is the screw strain vector, $\boldsymbol{\xi}^*$ is the reference screw strain, and $\Sigma$ is the cross-sectional screw stiffness matrix. Here, $\boldsymbol{M}$ and $\boldsymbol{K}$ are defined as in (4).

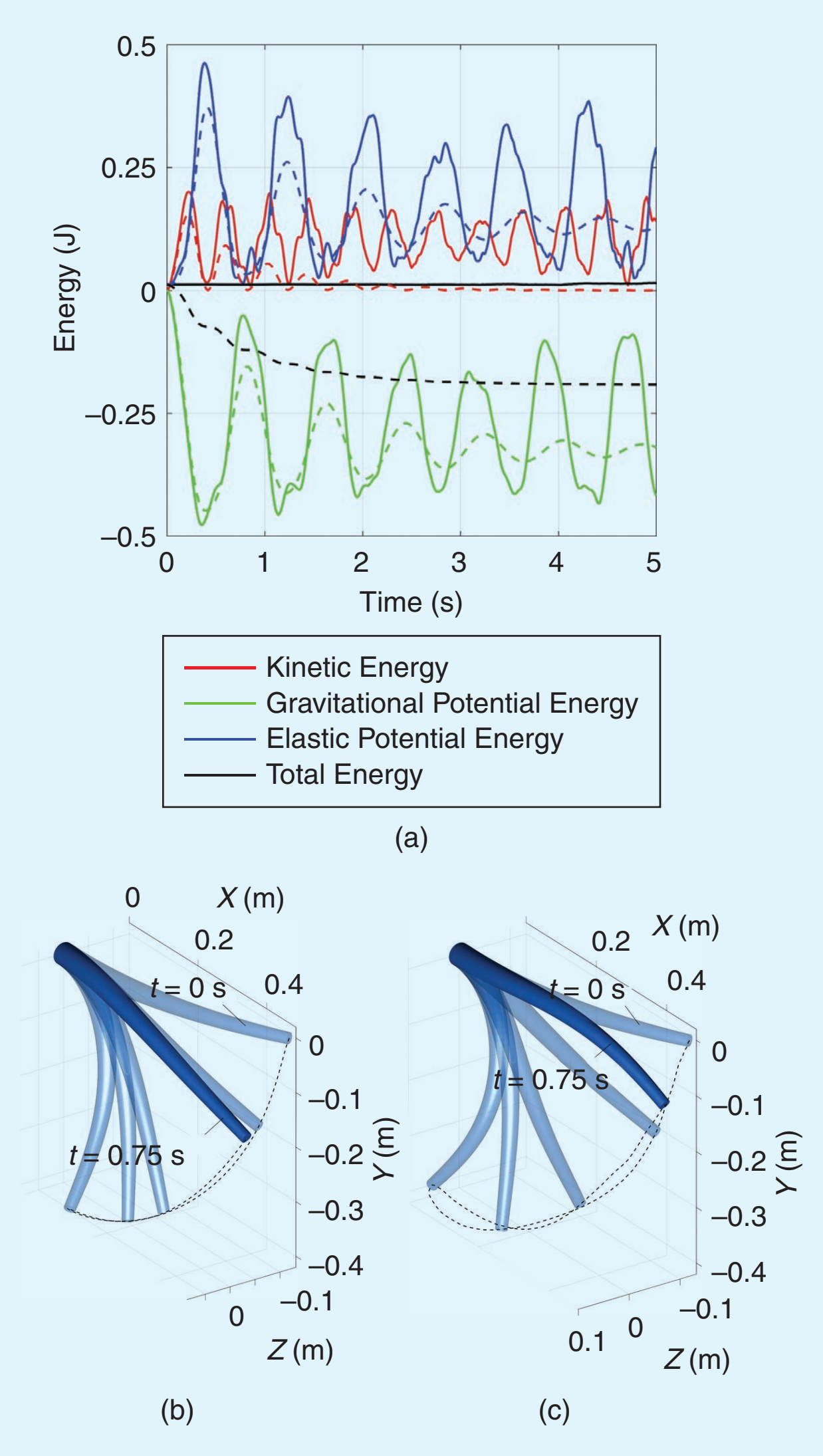

**Figure 7.** The free fall dynamics. (a) The energy change for the undamped (solid) and damped (dashed) beam. Superimposed images of (b) the damped case and (c) the undamped case. Dashed lines denote the tip trajectory.

The first, second, and third terms on the right-hand side of the equation represent kinetic, gravitational potential, and elastic potential energies, respectively. We plot these for the damped and the undamped cases in Figure 7(a). The total energy of the damped system is decaying, and that of the undamped system remains a nonzero positive constant corresponding to the initial strain energy. The results highlight the robustness of the SoRoSim dynamics, as any deviation in the energy conservation is solely attributed to errors in numerical integration (in time and space).

For the undamped case, the system's total energy should remain the same as the initial elastic potential energy, while for the damped case, the total energy should decrease monotonically. Figure 7(a) conveys the change in the values of various forms of energies of the system. The total energy of the damped system is decaying, and that of the undamped system remains a nonzero positive constant, as expected. This example also emphasizes that apart from dynamic and static simulations, we can use the toolbox for postprocessing. The methods of the SorosimLinkage class allow the user to compute quantities, such as the configuration (the position and orientations of cross sections), velocities, and generalized mass matrices, as functions of $\boldsymbol{q}$ and $\dot{\boldsymbol{q}}$. The user can compute these values in a separate MATLAB code for postprocessing, such as the energy computations in this example.

## Modeling Applications

This section presents examples of the static equilibrium and dynamic simulations relevant to the soft robotics field that the toolbox can solve. We demonstrate the applicability of the toolbox to different systems and the use of custom functions to model various external forces, including contact and fluid interactions.

### Hybrid Manipulators

Serial robotic arms are widely used to automate manufacturing processes and any task requiring object gripping and manipulation. We show the toolbox's ability to model and analyze hybrid manipulators for two different cases.

#### Case 1: Static Equilibrium

This example demonstrates the static equilibrium analysis of a hybrid (rigid-continuum) robotic arm with soft grippers under a wide range of loading conditions, including gravity (distributed load), a point force ($F$), rigid joint actuation, and cable actuation of soft links for gripping. We use the toolbox to investigate the static equilibrium analysis of the manipulator and demonstrate its ability in modeling systems with multiple types of links, joints, and forces within the same framework.

The system consists of 11 links that form open-, branched-, and closed-chain components, as illustrated in Figure 8(a). The first link is a rigid link that is connected to the ground via a planar joint, which allows displacement in the *xy*-plane and rotation about the *z*-axis. The second and third links are rigid links connected using revolute joints, which rotate about the *y*- and *z*-axes, respectively. These three rigid links are controlled by their joint coordinate values. The toolbox allows the user to enter values of their joint coordinates as inputs to the simulation, as shown in Figures 8(b)(2)–(4). Following the third link, there is a passive buffer consisting of three parallel soft links connected to two rigid disks at each end, as in Figure 8(c). These soft links are attached to the base disk via spherical joints and connected to the end disk via closed-loop fixed joints. The buffer, which adds to the system's complexity, is an example of a multi-DoF joint, parallel link, and closed-loop component, which the SoRoSim toolbox can model. We assign two divisions for these soft links, enable all rotational DoF, and assign constant strain bases. Using appropriate reference strain values, we set the precurvatures of the soft buffer links. An angle-controlled rigid disk attached at the end of the buffer assists further orientation of the grippers by rotating about its local $x$-axis as in Figure 8(b)(5).

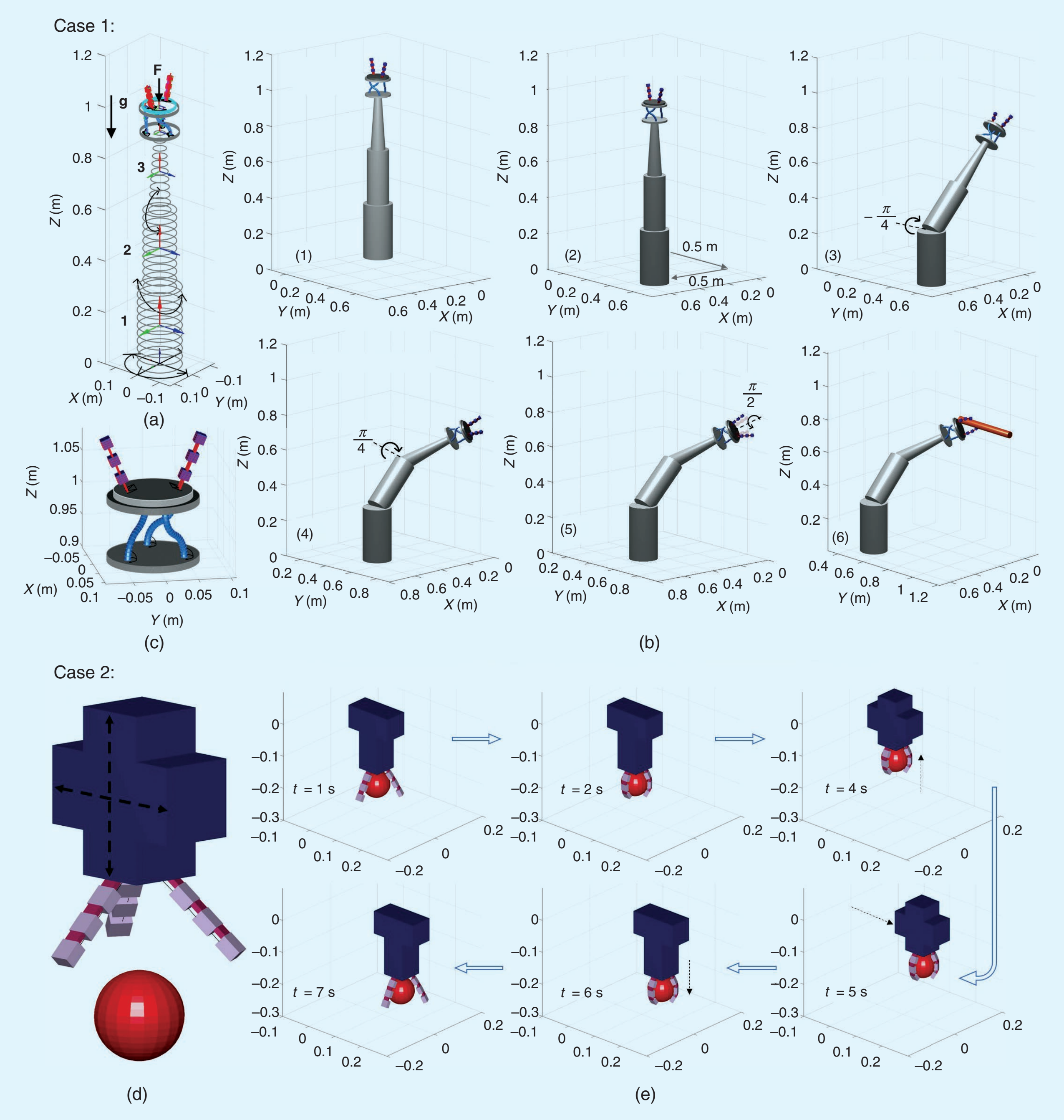

**Figure 8.** Case 1: (a) the hybrid robotic arm, showing some loads the system is subjected to; (b) the static equilibrium configurations of the arm at different stages of actuation input; and (c) a close-up view of the manipulator tip, showing the buffer with three parallel closed-loop soft links and the cable-actuated gripper fingers. The (1) initial position, (2) translation of link 1, (3) rotation of link 2, (4) rotation of link 3, (5) gripper orientation, and (6) object gripping. Case 2 (contact modeling): (d) the hybrid robot and grip target and (e) snapshots of the object manipulation procedure.

Finally, we have a soft gripper with two fingers that are cable actuated. The thin sections of the finger act as flexible links, while the thick sections act as rigid links. We assign a strain basis with constant bending about the local $y$-axis to the thin sections. This design of the gripper, where the cable is positioned outside the thin soft section, is widely used in the soft robotic community [13]. This particular mode of actuation, where the actuators go in and out of the system, is an important class of prototype that SoRoSim is able to simulate.

We compute the static equilibrium configuration of the system for the actuation inputs given in Figure 8(b). The system is subject to gravity and a follower point force. The direction of the gravity and the point force ($F$) are included in Figure 8(a). The value of the point force is 2 N. The corresponding equilibrium configurations are detailed in Figure 8(b)(2)–(5). Figure 8(b)(6) demonstrates the actuation of the grippers when a cable tension of 2 N is applied. We show a rod as a reference grip target. There are 39 DoF and seven actuation inputs for the system. On average, the static equilibrium simulations [shown in Figure 7(b)(1)–(6)] take 1.2 s.

This example analyzed a system with passive and actuated rigid joints controlled by joint coordinate values. The toolbox can also define joints controlled by joint force and moment values. The user can also assign stiffness values for rigid joints. For instance, a prismatic joint with a positive stiffness value is equivalent to a linear spring.

### Case 2: Contact Dynamics

Analysis of mechanical systems involving contact–impact events is demanding for many real-world applications. Contact dynamics involve determining potential contact points between colliding bodies, evaluating contact–impact forces, and establishing the transition between different contact scenarios. The implementation of contact mechanics is one of the most challenging and complex problems in modeling and computation [23].

Contact between two points involves the application of equal and opposite normal and tangential forces on each other. The current SoRoSim toolbox does not have built-in capability for contact dynamics. However, the user can apply various kinds of external loads by enabling a property of the SorosimLinkage class, namely, "Custom External Force Present," and then editing a MATLAB function, "CustomExtForce.m," to model the external force. Quantities such as Jacobian $\boldsymbol{J}$, screw velocity $\boldsymbol{\eta}$, and transformation matrices $\boldsymbol{g}$ are passed as inputs into the "CustomExtForce" function. The user can apply a custom external force as a function of these quantities.

To demonstrate contact dynamics using SoRoSim, we analyze a three-fingered gripper system attempting to grip and manipulate a spherical rigid body [Figure 8(d)]. The system consists of two rigid links with prismatic joints and three symmetrically arranged fingers similar to the previous case. We choose the dimensions of links arbitrarily. The inputs to the dynamic simulations are the kinematic inputs of the prismatic joints (vertical and horizontal translation) and the cable tension, which actuates the fingers.

We make the following approximations to simplify the estimation of contact points and the evaluation of contact forces.

1) The three thick rigid sections of each finger are approximated as spheres with diameters ($2r_i$) equal to the section heights and centered at the center of mass of the links.
2) We neglect the tangential (frictional) component of the contact forces. The normal force due to nine different contact points (three on each finger) is sufficient to provide a force closure grasp.
3) We use a linear spring with a linear damper to represent the normal contact force ($\boldsymbol{f}$). The contact force acting on the $i$th link is given by

$$\boldsymbol{f_i} = H(\delta_i) \cdot (k\delta_i + b\dot{\delta}_i)\frac{\boldsymbol{p}_{si}}{\| \boldsymbol{p}_{si} \|} \tag{6}$$

where $\boldsymbol{p}_{si}$ is the vector from the center of the sphere to the center of the $i$th link, $\delta_i = r_s + r_i - \| \boldsymbol{p}_{si} \|$, $r_s$ is the radius of the spherical target, $k$ is the contact stiffness constant, $b$ is a damping constant, and $H(\delta_i)$ is the Heaviside function, which ensures that the force is applied only if there is contact $(\delta_i > 0)$. While $\boldsymbol{f_i}$ acts on the $i$th link, $\Sigma \boldsymbol{f_i}$ acts on the grip target. Figure 8(e) contains snapshots of the grip manipulation maneuver.

While the gripper in Figure 1(f) is a planar equivalent of this example, the robots in Figure 1(a), (b), and (e) can be modeled as rigid–soft hybrid robots with contact dynamics. This example demonstrates that we can use the SoRoSim toolbox to analyze robotic systems with collision and contact events.

## *Underwater Robotics*

Due to their compliant nature, soft robots pose no risk to underwater habitats or organisms, making them an excellent choice to employ in underwater exploration. Bacteria-inspired flagellate propellers could be employed for the locomotion of robots underwater. The idea of soft propellers is based on the observation that an inclined rotation of a soft filament in a liquid environment generates a helical filament shape, which produces positive thrust for propulsion. An advanced version of an underwater robot propelled by four soft flagellate modules is presented in [14], with experimental validations of the dynamic simulation. For those simulations, the authors used custom-made MATLAB codes with constant strain (order 0 strain polynomials) approximations. Here, using the SoRoSim toolbox, we demonstrate the variable-strain dynamics of a similar system with a propeller consisting of three flagella [Figure 9(a)].

To rotate the filaments in an inclined fashion, we attach them to the terminals of a rigid body, namely, the hook [the inset in Figure 9(a)]. All hook terminals have an inclination of 45° with the local $x$-axis. They are also separated by a rotation of 120° with respect to one another. The hook is connected to a motor shaft that provides the required torque. Additional components, such as motors, electronics, the battery, and balancing weights, are kept inside the spheroidal shell of the robot. We create a SorosimLinkage consisting of six links: the shell, shaft, hook, and three filaments. The shell is modeled as a rigid link with an adjusted inertia matrix to account for the internal components and spheroidal (nondefault) shape. Its joint is a passive prismatic joint that allows translation only along the $x$-axis. The shaft is modeled as a cylindrical rigid link with an angle-controlled revolute joint to simulate the motor's input to the propeller. The hook is modeled as a fixed-joint rigid link with an adjusted inertia matrix to account for its shape. Finally, we define the three 0.7-m-long soft filaments as fixed-joint soft links with linear angular strains (torsion about the $x$-axis and bending about the $y$- and $z$-axes). The dimensions and material properties of these components are arbitrary but inspired by the proposed design in [14].

To simulate the fluid–robot interaction forces described in Figure 9(a), we need to include the external forces due to the presence of the fluid (buoyancy, drag, and lift forces) and the forces due to the volume of fluid moved by the propeller (added mass), described in [14], that are not part of the default force inputs during the SorosimLinkage creation. However, this does not prevent the user from implementing them using the "CustomExtForce.m" function.

By providing an actuation input of $\theta = 2\pi t$, we run the simulation to capture the system's dynamic response for the first 5 s. Figure 9(b) presents superimposed images of the motion of the robot, starting from rest to $t = 1.5$ s. In the inset, we plot the robot's forward velocity as a function of time. The combination of actuation and environmental interaction leads to the emergence of intelligent helical deformations of the filaments, which, in turn, cause a thrust that propels the robot forward. This example emphasizes the usefulness of the SoRoSim toolbox in modeling rigid–soft hybrid underwater robots. The toolbox can model the robots mentioned Figure 1(c) and (g) in a similar way.

## Design Analysis and Control Applications

In addition to static equilibrium analysis and dynamic simulations, we can use the toolbox for design optimization, inverse kinematics, and control applications. In this section, we demonstrate examples of an optimization problem and two control problems that can be solved using the SoRoSim package and custom-made MATLAB codes.

### Design Optimization

Here, we demonstrate the use of the toolbox in optimizing an objective function based on the static equilibrium of a single cable-actuated soft manipulator. We begin by defining a soft link with two divisions, with a total length of 80 cm and a radius linearly varying from 4 to 2 cm. We use a Young's modulus of 10 MPa and a Poisson's ratio of 0.5. We enable all the rotational DoF (torsion about the $x$-axis and bending about the $y$- and $z$-axes) and approximate the corresponding strains using a first-order polynomial. No external forces are taken into account for the static equilibrium. We set up an optimization problem to find an appropriate cable path and cable tension, minimizing the error in the manipulator end, and middle point coordinates with desired positions at static equilibrium. The objective function is given by

$$\Omega = \| \bar{\boldsymbol{r}}_{\mathrm{mid}} - \boldsymbol{r}_{\mathrm{mid}} \| + \| \bar{\boldsymbol{r}}_{\mathrm{end}} - \boldsymbol{r}_{\mathrm{end}} \| \tag{7}$$

where $\boldsymbol{r}_{\mathrm{mid}}$ and $\boldsymbol{r}_{\mathrm{end}}$ are the equilibrium positions of the midpoint and the endpoint of the manipulator and $\bar{\boldsymbol{r}}_{\mathrm{mid}}$ and $\bar{\boldsymbol{r}}_{\mathrm{end}}$ are the desired mid- and endpoint positions.

We set the desired midpoint $\bar{\boldsymbol{r}}_{\mathrm{mid}}$ to $(0.35, -0.1, -0.1)$ and the desired tip point $\bar{\boldsymbol{r}}_{\mathrm{end}}$ to $(0.45, -0.1, -0.4)$. Since the cable position is constrained to be inside the soft manipulator, the problem is a bounded optimization problem with an inequality condition. We parametrize the cable path's $y$ and $z$ coordinates by using quadratic polynomials. Hence, we need to optimize seven variables: the coefficients of the quadratic polynomials defining $y_c$ and $z_c$ (the $y$ and $z$ coordinates of the cable as a function of $X$) and the cable tension that is applied.

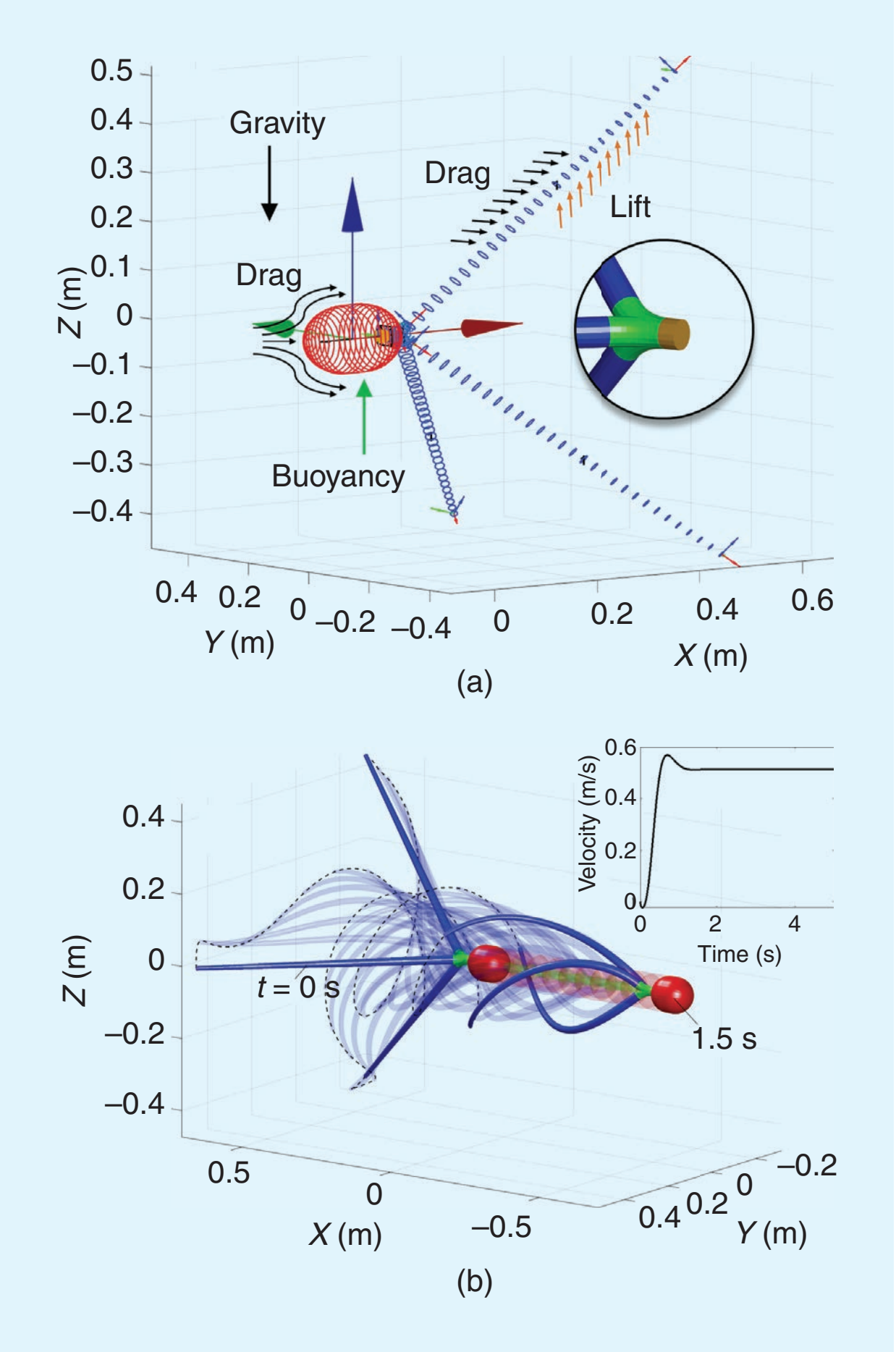

**Figure 9.** Underwater robotics. (a) The forces acting on the robot, with an inset showing the hook geometry. (b) The motion of the flagellate robot (superimposed images) due to the thrust generated by the soft propeller, with an inset showing the robot's speed versus time. Dotted lines show the tip trajectory of each soft filament.

We use the MATLAB "patternsearch" algorithm to find the optimal values of variables at the local minimum of the objective function. The initial condition is obtained from $y_c = 0X^2 + 0X + 0.02$, $z_c = 0X^2 + 0X + 0$, and $T = 0$ N. The "patternsearch" algorithm, with parallel computing enabled, takes about 1.5 min (93 s) to converge at the optimized parameters. There are 2,887 objective function evaluations (>30 evaluations per second), including static equilibrium and forward kinematics evaluation to estimate the mid and endpoint coordinates. The optimal solution is given by $y_c = 0.024X^2 + 0.003X - 0.010$, $z_c = 0.015X^2 - 0.012X - 0.012$, and $T = 112.5$ N. The initial and optimized cable path is shown in Figure 10(a) and (b). The static equilibrium state of the manipulator at the optimized parameters is in Figure 10(c). The manipulator tip position matches perfectly with the desired tip position. However, there is an error of about 14% in the midpoint position. The error could be due to geometric constraints, such as the length and radius of the manipulator. It may be decreased by testing different initial values for the optimization.

This example stresses that the user can combine SoRoSim with well-established MATLAB packages, such as the global optimization toolbox, to deliver analysis results outside the package of the SoRoSim toolbox. The user may also define an optimization problem based on a dynamic simulation to estimate the design parameters of a multibody system that optimizes its dynamic performance.

### Inverse Dynamic Control

Finally, we use the toolbox to solve two inverse dynamics control problems. We begin by creating a soft link that is 1 m long, consisting of two divisions (0.5 m each). The radius of the link varies linearly from 2 to 1 cm from the base to the tip [Figure 11(a)]. The material properties are the same as those used for the optimization example. For the first division, we define a linear bending about the $y$-axis and the $z$-axis. In the second section, the same deformation modes are defined using a constant strain basis. The soft manipulator is actuated using six linearly independent cables (Table 1).

The velocity of the manipulator tip is given by $\boldsymbol{\eta}_t = \boldsymbol{J}_t\dot{\boldsymbol{q}}$, where $\boldsymbol{\eta}_t$ is the tip velocity and $\boldsymbol{J}_t$ is the Jacobian at the tip. Taking a time derivative of the equation and substituting $\ddot{\boldsymbol{q}}$ from (4), we get the dynamics equation of the manipulator tip. From this equation, if we solve for the actuator strength, $\boldsymbol{u}$, and apply a proportional derivative (PD) controller [24], we get the following task space control law:

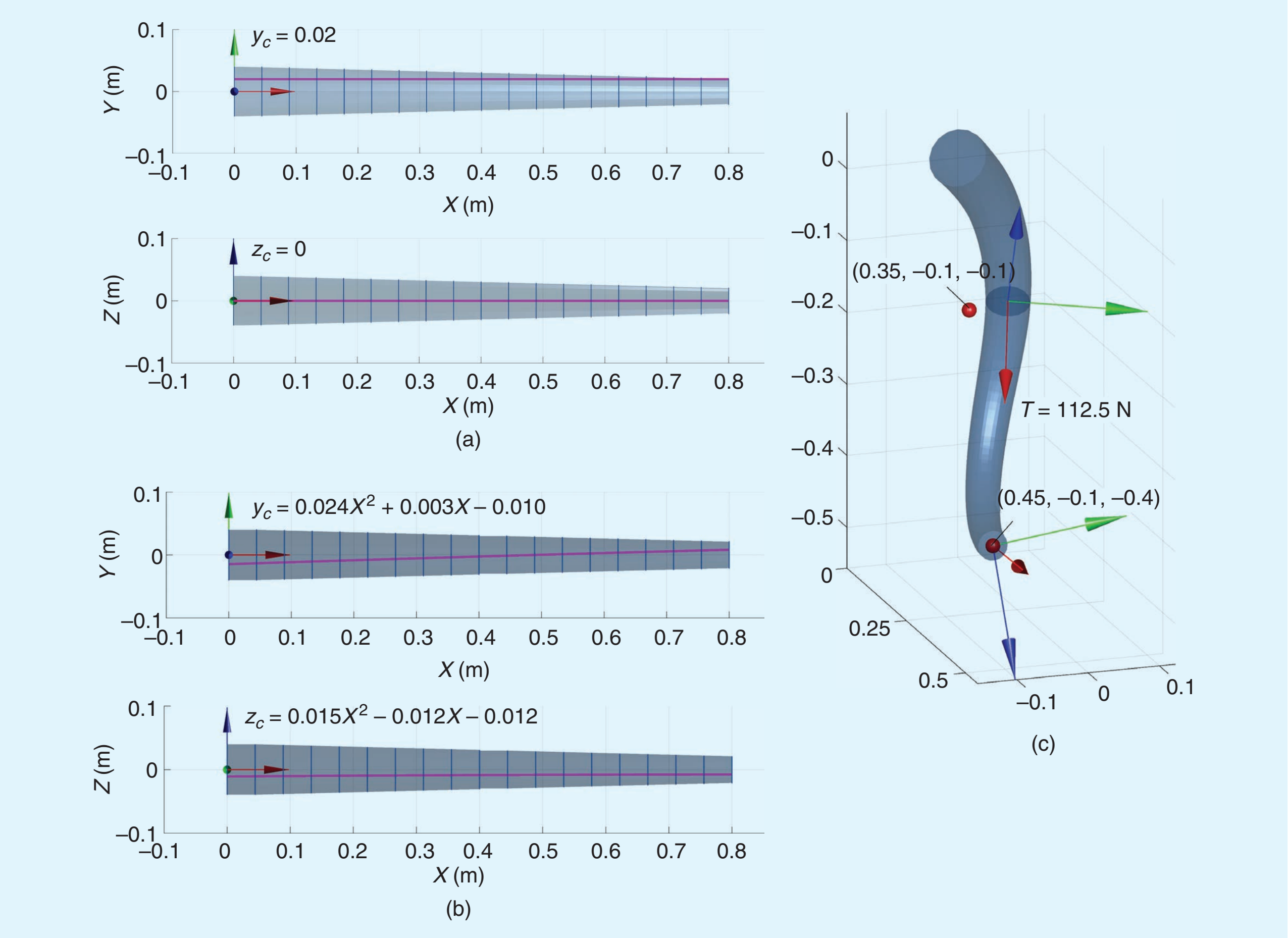

**Figure 10.** The (a) initial and (b) optimized cable paths. (c) The final configuration, where the optimized cable path and tension are used. Red points indicate the desired points.

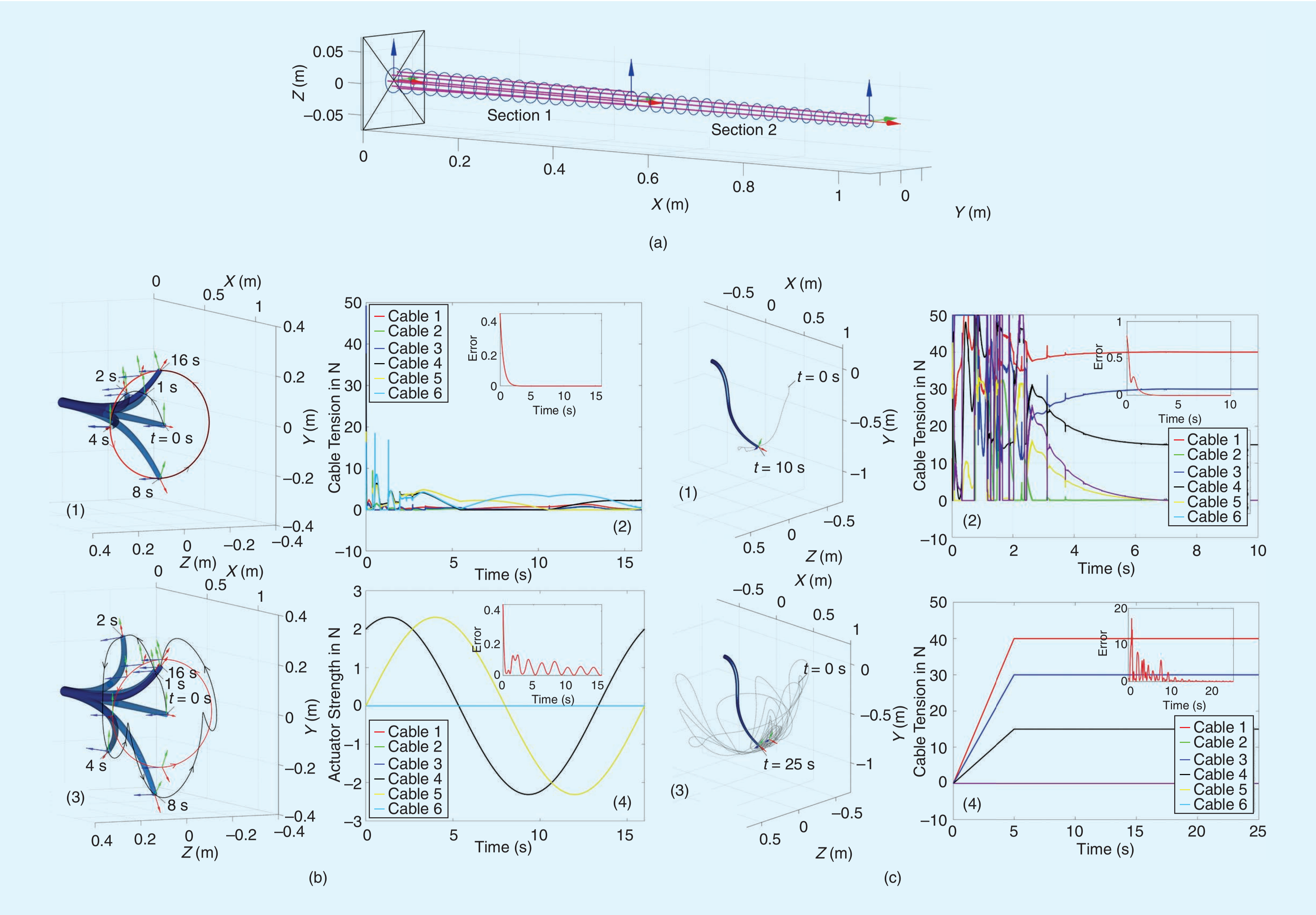

**Figure 11.** The control of a soft manipulator. (a) The definition of the manipulator with six actuators. (b) The circular trajectory control. (c) The fixed position and orientation control under gravity. Here, (1) and (2) are the outputs of the closed-loop proportional derivative controller, while (3) and (4) are those of an open-loop input.

$$u = (J_t M^{-1} B)^{\dagger}[\dot{\bar{\eta}}_t + K_d(\bar{\eta}_t - \eta_t) + K_p(\log(g_t^{-1}\bar{g}_t))^{\vee} + J_t M^{-1}((C+D)\dot{q} + Kq + Q) - \dot{J}_t\dot{q}] \quad (8)$$

where $K_p$ and $K_d$ are the proportional and derivative gains, $\dot{\bar{\eta}}_t$ is the desired tip acceleration, $\bar{\eta}_t$ is the desired tip velocity, $g_t$ and $\bar{g}_t$ are the actual and desired tip transformation matrices, and *log* is the logarithmic operator in *SE*(3). Note that the proposed control law may not ensure a full state ($q$ and $\dot{q}$) convergence of the manipulator at the steady state for an underactuated system [24]. Addressing issues of task space-based controllers is out of this article's scope. Here, we demonstrate the toolbox's ability to incorporate the control law.

The user can define a custom actuator strength by enabling the "Custom Actuator Strength" property of the SorosimLinkage and by editing the "CustomActuator Strength.m" file. Quantities such as Jacobian $J$, derivatives of Jacobian $\dot{J}$, and generalized mass matrix $M$ are passed as inputs into the "CustomActuatorStrength" function. The user can directly use these to compute the actuator strength given by (8).

### Case 1: Tip Pose Trajectory Tracking

Here, we attempt to control the position and orientation of the manipulator tip based on those of a reference frame moving in a circular trajectory. The circular trajectory we define has a radius of 0.22 m, parallel to the *yz*-plane, and its center is located at 0.96 m on the *x*-axis. Moreover, the reference frame of the trajectory is at 35.96° with respect to the *x*-axis. The angular velocity of the reference frame is set to 3.75 r/min. We run the dynamic simulation for 16 s, using the actuator strength computed using (8). Superimposed images of the dynamic results, corresponding actuator strengths, and errors (inset) are given in Figure 11(b)(1) and (2). The dynamics of the system are perfectly cancelled out by the controller, and the error between the tip and the reference frames converges to zero in fewer than 4 s. The plot in Figure 11(b)(2) details the process of nullifying the system dynamics at the beginning and the oscillation of the actuator strength as time elapses. We use the "lsqlin" function of MATLAB to compute the actuator strength given by (8). Using this function, we ensure that the cable tension is a positive value of less than 50 N.

To compare the controller's performance, we input the actuator strength corresponding to the quasi-static solution that matches the position and orientation of the manipulator tip and the moving frame. The dynamics of the simulation are available in Figure 11(b)(3) and (4). In this case, the error decreases initially but continues to oscillate indefinitely due to the dynamic response of the system.

**Table 1. The cable coordinates ($d_1$ = 1.5 cm and $d_2$ = 1 cm).**

| Cable Number | Cable Length (m) | *y* Coordinate | *z* Coordinate |
|---|---|---|---|
| 1 | 0.5 | $\frac{1}{2}d_1$ | $\frac{\sqrt{3}}{2}d_1$ |
| 2 | 0.5 | $-d_1$ | 0 |
| 3 | 0.5 | $\frac{1}{2}d_1(1-X)$ | $-\frac{\sqrt{3}}{2}d_1(1-X)$ |
| 4 | 1 | $d_2$ | 0 |
| 5 | 1 | $-\frac{1}{2}d_2$ | $\frac{\sqrt{3}}{2}d_2$ |
| 6 | 1 | $-\frac{1}{2}d_2\left(1-\frac{X}{2}\right)$ | $\frac{\sqrt{3}}{2}d_2\left(1-\frac{X}{2}\right)$ |

### Case 2: Tip Pose Regulation Under Gravity

For the second control case, we attempt to control the position and orientation of the manipulator tip based on a fixed reference frame under the influence of gravity (external force). The fixed reference frame is the same as the manipulator tip frame at the static equilibrium when the cable tensions are 40, 0, 30, 15, 0 , and 0 N, respectively, for each cable.

Using the PD controller given by (8), we get a dynamic response, as shown in Figure 11(c)(1) and (2). We can see a steady approach to the desired tip position and orientation, denoted by the reference frame in the figure. The controller overcomes the dynamic response due to gravity and the actuation force, and the error is quickly reduced to zero in fewer than 5 s. Over time, the cable tensions converge to constant values corresponding to the static equilibrium case. To compare the controller performance, we input cable tensions as ramp functions that increase to reach the static equilibrium cable tension values at 5 s. This manual actuator strength input is unable to quickly cancel the system's oscillatory response [Figure 11(c)(3)]. The error is observed to decrease slowly, allowing the tip to approach the desired position and orientation.

## Discussion and Conclusion

The computational performance of the toolbox depends on problem parameters, such as the number of links, the DoF of the system, strain order used, number of Gauss quadrature points on soft links, number of actuators, values of actuator strengths, external forces, material, and geometric properties. Here, we report the simulation speed of the current version of the SoRoSim toolbox. Table 2 summarizes parameters that characterize the computational performance of the toolbox. For each analysis discussed in the "Toolbox Validation" and "Modeling Applications" sections, the table shows information, including the total number of links ($N$), total DoF ($N_{\mathrm{dof}}$), maximum strain order [$\xi(O)$], order of the Zannah collocation approximation [$Z(O)$], and total number of points at which static or dynamic (4) equations are evaluated ($N_{\mathrm{eva}}$). For static analyses (highlighted in gray), the seventh column of the table indicates the total static equilibrium simulations performed ($N_s$). For dynamic simulations, the column shows the real-time duration (0 to $t_{\max}$) of the dynamics problem.

MATLAB solvers used to solve the static or dynamic equations are shown in the eighth column. The final column

**Table 2. A summary of the toolbox performance.**

| Example | $N$ | $N_{dof}$ | $\xi(O)$ | $Z(O)$ | $N_{eva}$ | $N_s$ (Number)/ $t_{max}$ (s) | Solver | $t_s$ (s) |
|---|---|---|---|---|---|---|---|---|
| "Test 1: Fixed–Free Beam With a Follower Tip Force" section | 1 | 5 | 4 | 4 | 16 | 130 | *fsolve* | 5.5 |
| "Test 2: FEM Study of a Fixed–Fixed L-Shaped Beam" section | 2 | 30 | 2 | 4 | 20 | 1 | *fsolve* | 0.5 |
| "Test 3: Dynamics of a Flexible Flying Rod" section | 1 | 15 | 2 | 4 | 13 | 7 | *ode*45 | 1 |
| "Test 4: Energy Balance in SoRoSim Dynamics" section | 1 | 12 | 3 | 4 | 15 | 5 | *ode*15*s* | 1.5 |
| "Test 4: Energy Balance in SoRoSim Dynamics" section* | 1 | 12 | 3 | 2 | 15 | 5 | *ode*113 | 6 |
| "Case 1: Static Equilibrium" section | 11 | 39 | 0 | 4 | 143 | 5 | *fsolve* | 6 |
| "Case 2: Contact Dynamics" section | 6 | 21 | 0 | 4 | 96 | 9 | *ode*15*s* | 27 |
| "Underwater Robotics" section | 6 | 20 | 1 | 2 | 30 | 5 | *ode*113 | 5 |
| "Design Optimization" section | 1 | 12 | 1 | 4 | 15 | 2,887 | *fsolve*, *patternsearch* | 93 |
| "Case 1: Tip Pose Trajectory Tracking" section | 1 | 6 | 1 | 4 | 15 | 16 | *ode*1† | 5 |
| "Case 2: Tip Pose Regulation Under Gravity" section | 1 | 6 | 1 | 4 | 15 | 10 | *ode*1† | 4 |

Static simulations are highlighted in gray.
*With undamped ($\boldsymbol{D} = \mathbf{0}$) dynamics.
†With a time step of 0.01 s.

shows the total simulation runtime ($t_s$). We used a PC with the following specifications for these analyses: Intel Core i7-1065G7 CPU at 1.3 GHz and 1.5 GHz and with 16 GB of random-access memory. The MATLAB version used for these analyses is R2021a. The comparison of $t_{\max}$ and $t_s$ indicates that the toolbox performs in real time or faster for most of the examples presented in this article. This is the result of the geometrically exact formulation, which requires the fewest DoF to estimate the state of a robot. Fast (2.5× faster than real time) computation of actuator strengths could be used for model-based inverse dynamic control of real-world soft manipulators. We will implement an implicit time integrator to speed up the simulation in a future version of the toolbox. Further code optimization and implementation of parallel computing can also increase the toolbox performance.

It is important to emphasize that, unlike FEM packages, the toolbox is suitable only for systems whose soft links can be modeled as Cosserat rods. Despite this limitation, the geometrically exact approach used in the toolbox makes it a fast (due to a smaller number of DoFs) and accurate tool for systems involving large deformations, which is the case for the majority of soft robotics applications. The toolbox allows the modeling and analysis of systems with open-, branched-, and closed-chain and interconnected structures. Additionally, the toolbox provides the user with plenty of ways to use the output data with existing MATLAB packages and user-written MATLAB codes.

In summary, this article presented SoRoSim, an intuitive MATLAB toolbox that uses the GVS approach to provide a unified framework for the modeling, analysis, and control of soft, rigid, and hybrid robots. It provides users with a level of freedom that will enable them to tailor and adjust material models, external forces, actuation paths, and functions to fit their applications and intended systems. We validated the toolbox simulation results by comparing the analysis results with existing literature and numerical studies. We provided four examples to highlight the application of the toolbox in soft robotics. SoRoSim successfully bridges the gap between soft and traditional robotics modeling and analysis. It is an ongoing effort and will be under constant improvement in terms of performance, features, and GUI enhancement. We hope that the users of SoRoSim find it beneficial and user friendly, and we look forward to it being widely used in the research and engineering community.

## Acknowledgment

This work was supported, in part, by the U.S. Office of Naval Research Global, under grant N62909-21-1-2033, and Khalifa University of Science and Technology, under grants CIRA-2020-074 and RC1-2018-KUCARS. Anup Teejo Mathew and Ikhlas Ben Hmida are co-first authors. This article contains supplementary material available at https://doi.org/10.1109/MRA.2022.3202488, provided by the authors.

## References

[1] C. Laschi, B. Mazzolai, and M. Cianchetti, "Soft robotics: Technologies and systems pushing the boundaries of robot abilities," *Sci. Robot.*, vol. 1, no. 1, p. eaah3690, Dec. 2016, doi: 10.1126/scirobotics.aah3690.

[2] C. Armanini, C. Messer, A. T. Mathew, F. Boyer, C. Duriez, and F. Renda, "Soft robots modeling: A literature unwinding," 2021. [Online]. Available: https://arxiv.org/abs/2112.03645

[3] D. Holland, S. Berndt, M. Herman, and C. Walsh, "Growing the soft robotics community through knowledge-sharing initiatives," *Soft Robot.*, vol. 5, no. 2, pp. 119–121, Apr. 2018, doi: 10.1089/soro.2018.29013.dph.

[4] M. A. Graule, C. B. Teeple, T. P. McCarthy, R. C. St. Louis, G. R. Kim, and R. J. Wood, "SOMO: Fast and accurate simulations of continuum

robots in complex environments," in *Proc. 2021 IEEE Int. Conf. Intell. Robots Syst. (IROS)*, pp. 3934–3941, doi: 10.1109/IROS51168.2021.9636059.

[5] F. Faure et al., "SOFA: A multi-model framework for interactive physical simulation," in *Soft Tissue Biomechanical Modeling for Computer Assisted Surgery, ser. Studies in Mechanobiology, Tissue Engineering and Biomaterials*, vol. 11, Y. Payan, Ed. Berlin, Germany: Springer-Verlag, Jun. 2012, pp. 283–321. [Online]. Available: https://hal.inria.fr/hal-00681539

[6] Y. Hu et al., "ChainQueen: A real-time differentiable physical simulator for soft robotics," in *Proc. 2019 Int. Conf. Robot. Automat.*, pp. 6265–6271, doi: 10.1109/ICRA.2019.8794333.

[7] S. Grazioso, G. D. Gironimo, and B. Siciliano, "A geometrically exact model for soft continuum robots: The finite element deformation space formulation," *Soft Robot.*, vol. 6, no. 6, pp. 790–811, Dec. 2019, doi: 10.1089/soro.2018.0047.

[8] J. Austin, R. Corrales-Fatou, S. Wyetzner, and H. Lipson, "Titan: A parallel asynchronous library for multi-agent and soft-body robotics using NVIDIA CUDA," in *Proc. 2020 Proc. IEEE Int. Conf. Robot. Automat. (ICRA)*, pp. 7754–7760, doi: 10.1109/ICRA40945.2020.9196808.

[9] N. Naughton, J. Sun, A. Tekinalp, T. Parthasarathy, G. Chowdhary, and M. Gazzola, "Elastica: A compliant mechanics environment for soft robotic control," *IEEE Robot. Automat. Lett.*, vol. 6, no. 2, pp. 3389–3396, Apr. 2021. [Online]. Available: https://ieeexplore.ieee.org/document/9369003, doi: 10.1109/LRA.2021.3063698.

[10] B. Angles et al., "Viper: Volume invariant position-based elastic rods," *Proc. ACM Comput. Graph. Interact. Techn.*, vol. 2, no. 2, pp. 1–26, Jun. 2019, doi: 10.1145/3340260.

[11] S. H. Sadati et al., "TMTDyn: A Matlab package for modeling and control of hybrid rigid–continuum robots based on discretized lumped systems and reduced-order models," *Int. J. Robot. Res.*, vol. 40, no. 1, pp. 296–347, Jan. 2021, doi: 10.1177/0278364919881685.

[12] C. Armanini, I. Hussain, M. Z. Iqbal, D. Gan, D. Prattichizzo, and F. Renda, "Discrete cosserat approach for closed-chain soft robots: Application to the fin-ray finger," *IEEE Trans. Robot.*, vol. 37, no. 6, pp. 1–10, Dec. 2021. [Online]. Available: https://ieeexplore.ieee.org/abstract/document/9442856, doi: 10.1109/TRO.2021.3075643.

[13] I. Hussain et al., "Modeling and prototyping of an underactuated gripper exploiting joint compliance and modularity," *IEEE Robot. Automat. Lett.*, vol. 3, no. 4, pp. 2854–2861, Oct. 2018. [Online]. Available: https://ieeexplore.ieee.org/abstract/document/8378053, doi: 10.1109/LRA.2018.2845906.

[14] C. Armanini, M. Farman, M. Calisti, F. Giorgio-Serchi, C. Stefanini, and F. Renda, "Flagellate underwater robotics at macroscale: Design, modeling, and characterization," *IEEE Trans. Robot.*, vol. 38, no. 2, pp. 731–747, Apr. 2022. [Online]. Available: https://ieeexplore.ieee.org/document/9486942, doi: 10.1109/TRO.2021.3094051.

[15] "SOROSIM: A unified framework for soft and rigid robotics simulation, control and optimization." GitHub. Accessed: Jul. 11, 2021. [Online]. Available: https://github.com/Ikhlas-Ben-Hmida/SoRoSim

[16] F. Renda, C. Armanini, V. Lebastard, F. Candelier, and F. Boyer, "A geometric variable-strain approach for static modeling of soft manipulators with tendon and fluidic actuation," *IEEE Robot. Automat. Lett.*, vol. 5, no. 3, pp. 4006–4013, Jul. 2020. [Online]. Available: https://ieeexplore.ieee.org/document/9057619, doi: 10.1109/LRA.2020.2985620.

[17] F. Boyer, V. Lebastard, F. Candelier, and F. Renda, "Dynamics of continuum and soft robots: A strain parameterization based approach," *IEEE Trans. Robot.*, vol. 37, no. 3, pp. 847–863, Jun. 2021, doi: 10.1109/TRO.2020.3036618.

[18] F. Renda and L. Seneviratne, "A geometric and unified approach for modeling soft-rigid multi-body systems with lumped and distributed degrees of freedom," in *Proc. IEEE Int. Conf. Robot. Automat. (ICRA)*, May 2018, pp. 1567–1574, doi: 10.1109/ICRA.2018.8461186.

[19] F. Boyer and F. Renda, "Poincare's equations for cosserat media: Application to shells," *J. Nonlinear Sci.*, vol. 27, no. 1, pp. 1–44, Feb. 2017, doi: 10.1007/s00332-016-9324-7.

[20] J. Selig, *Geometric Fundamentals of Robotics*. New York, NY, USA: Springer-Verlag, 2007. [Online]. Available: https://link.springer.com/book/10.1007/b138859

[21] J. C. Simo and L. Vu-Quoc, "A three-dimensional finite-strain rod model. Part II: Computational aspects," *Comput. Methods Appl. Mechanics Eng.*, vol. 58, no. 1, pp. 79–116, Oct. 1986, doi: 10.1016/0045-7825(86)90079-4.

[22] J. Simo and L. Vu-Quoc, "On the dynamics in space of rods undergoing large motions - A geometrically exact approach," *Comput. Methods Appl. Mechanics Eng.*, vol. 66, no. 2, pp. 125–161, Feb. 1988, doi: 10.1016/0045-7825(88)90073-4.

[23] P. Flores, "Contact mechanics for dynamical systems: A comprehensive review," *Multibody Syst. Dyn.*, vol. 54, no. 2, pp. 127–177, Feb. 2022, doi: 10.1007/s11044-021-09803-y.

[24] C. D. Santina, C. Duriez, and D. Rus, "Model based control of soft robots: A survey of the state of the art and open challenges," 2021. [Online]. Available: https://arxiv.org/abs/2110.01358

***Anup Teejo Mathew***, Department of Mechanical Engineering, Khalifa University of Science and Technology, Abu Dhabi 127788, United Arab Emirates. E-mail: anup.mathew@ku.ac.ae.

***Ikhlas Ben Hmida***, Department of Mechanical Engineering, Khalifa University of Science and Technology, Abu Dhabi 127788, United Araa Emirates. E-mail: ikhlas.benhmida@ku.ac.ae.

***Costanza Armanini***, Department of Mechanical Engineering, Khalifa University of Science and Technology, Abu Dhabi 127788, United Arab Emirates. E-mail: costanza.armanini@ku.ac.ae.

***Frederic Boyer***, Laboratory of Digital Sciences of Nantes, IMT Atlantique, Nantes 44307 France. E-mail: frederic.boyer@imt-atlantique.fr.

***Federico Renda***, Department of Mechanical Engineering, Khalifa University of Science and Technology, Abu Dhabi 127788, United Arab Emirates, and Khalifa University Center for Autonomous Robotic Systems, Khalifa University, Abu Dhabi 127788, United Arab Emirates. E-mail: federico.renda@ku.ac.ae.